\documentclass[runningheads]{llncs}
\usepackage[FINAL,year=2026,ID=6129]{eccv}

\usepackage{eccvabbrv}
\usepackage{multirow}
\usepackage{graphicx}
\usepackage{booktabs}

\usepackage[accsupp]{axessibility}

\usepackage{hyperref}
\usepackage{orcidlink}
\usepackage{colortbl}
\usepackage{xcolor}
\usepackage{amssymb}
\usepackage{amsmath}
\usepackage{array}

\definecolor{tablegray}{gray}{0.9}

\newcommand{\equalcontrib}{\textsuperscript{$\dagger$}}
\newcommand{\corresponding}{\textsuperscript{$\boxtimes$}}
\begin{document}


\title{
WCog-VLA: A Dual-Level World-Cognitive Vision-Language-Action Model for End-to-End Autonomous Driving
}

\titlerunning{
WCog-VLA
}

\author{
Xuerun Yan\inst{1,2}\equalcontrib\and
Zhexi Lian\inst{1}\equalcontrib\and
Nuoheng Zhang\inst{1}\and
Shiyu Fang\inst{1}\and
Haoran Wang\inst{1}\and
Chen Lv\inst{2}\and
Jia Hu\inst{1}\corresponding\and
Binyang Song\inst{2}\corresponding
}

\authorrunning{
X. Yan et al.
}

\institute{
Tongji University, China \and
Nanyang Technological University, Singapore
\\[2mm]
\textsuperscript{$\dagger$} Equal contribution
\quad
\textsuperscript{$\boxtimes$} Corresponding author
}

\maketitle

\begin{abstract}
  Vision-Language-Action (VLA) models have advanced end-to-end autonomous driving. However, existing methods either lack comprehensive world cognition or suffer from fragmented world foresight, inherently confining these models to reactive driving. To address this limitation, we propose WCog-VLA, a novel dual-level \textbf{W}orld-\textbf{Cog}nitive \textbf{VLA} framework that successfully bridges semantic world forecasting with generative world evolution to achieve proactive autonomous driving. At the semantic level, WCog-VLA unifies world cognition and reasoning by incorporating 3D spatial perception and injecting agent tokens to capture the world dynamics, while concurrently enabling Game-theoretic Chain-of-Thought (Game-CoT) reasoning. At the generative level, we introduce the Aligned Decoupled Diffusion Transformer (ADDT) as a powerful generative world model that synthesizes physically-plausible joint multi-agent trajectories. Through scene representation alignment, ADDT reduces the number of denoising steps required and thus significantly accelerates inference. To facilitate strategic reasoning, we further construct a large-scale dataset featuring 85k Game-CoT annotations. Extensive experiments on the NAVSIM benchmark demonstrate that WCog-VLA achieves a State-Of-The-Art (SOTA) PDMS score of 92.9.
\end{abstract}

\section{Introduction}
\label{sec:intro}
End-to-end (E2E) autonomous driving has emerged as a dominant paradigm by directly mapping raw sensory inputs to planned trajectories \cite{jiang2023vad,hu2022st,hu2023planning,chen2024vadv2} within a unified and differentiable framework. Although these E2E models show remarkable performance in common scenarios, they often struggle in complex or long-tail situations \cite{chen2024end,zhou2025opendrivevla}. This fragility arises from insufficient causal reasoning and world knowledge, leaving the models unable to fully understand and reason about the surrounding environments.

\begin{figure}[tb]
  \centering
  \includegraphics[trim={1.1cm 6.85cm 1.1cm 1cm}, clip, width=0.90\textwidth]{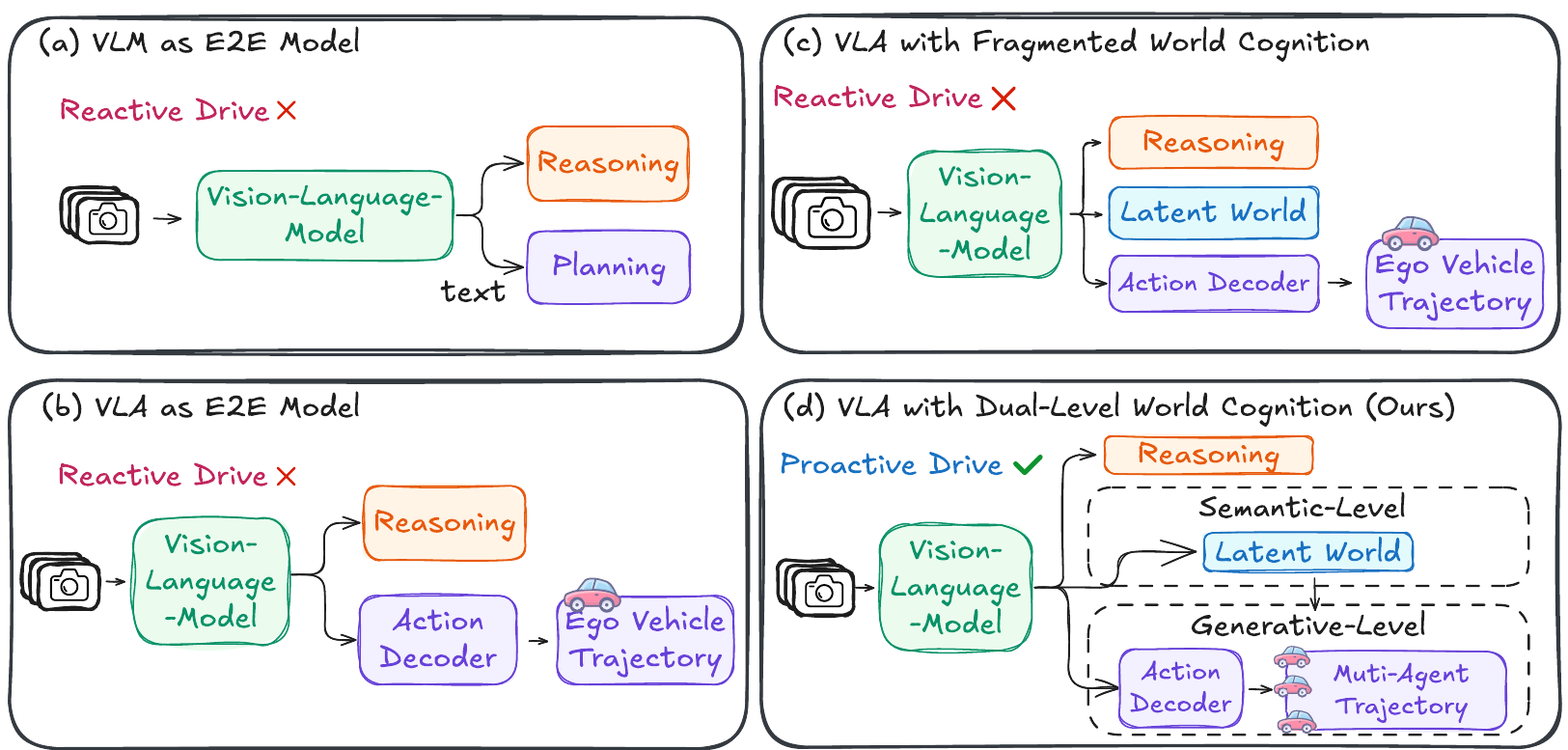}
  \caption{Four paradigms of leveraging VLM in E2E autonomous driving. Our method (d) advances existing frameworks to enable proactive driving by establishing a dual-level world cognition with the integration of semantic forecasting and generative evolution.}
  \label{fig:ECCV_intro}
  \vspace{-0.75em}
\end{figure}

To address these long-tail challenges, Vision-Language-Models (VLMs) \cite{achiam2023gpt,bai2025qwen3} have been increasingly integrated into E2E autonomous driving frameworks \cite{jiang2024senna,tian2024drivevlm,xing2025openemma}. Equipped with extensive world knowledge and strong reasoning capabilities, VLMs significantly advance scene comprehension in complex driving scenarios \cite{jiang2025survey,zhang2025pure}. Building upon this foundation, Vision-Language-Action (VLA) models further extend VLM capabilities to action generation. Existing VLA approaches typically operate in two ways. The first formulates action outputs as autoregressive sequence generation (\cref{fig:ECCV_intro}(a)), producing either discrete text tokens \cite{liu2025omnireason,luo2025adathinkdrive,qian2025agentthink} or learned action codes \cite{zhou2025autovla}. Alternatively, the second utilizes VLMs as cognitive encoders and attaches dedicated action decoders (\cref{fig:ECCV_intro}(b)) to generate continuous trajectories \cite{fu2025orion,xie2026latentvla}, \eg, diffusion models \cite{li2025recogdrive,jiang2025diffvla}.

Despite these advancements, existing VLA models still face several challenges:
(1) \emph{Lack of 3D spatial awareness.} Relying primarily on 2D image features, current models lack structured 3D spatial representations of surrounding road participants \cite{zhou2025opendrivevla,liu2025drivepi}, which are essential for accurate spatial reasoning and precise ego planning.
(2) \emph{Insufficient world cognition.} Existing methods struggle to adequately represent world states and forecast future dynamics \cite{team2026advancing}, such as the intentions of surrounding agents. This confines current methods to reactive rather than proactive driving (\cref{fig:ECCV_intro}(a),(b)). 
While some works incorporate world cognition via VLM hidden states \cite{li2026sgdrive, xiong2026unidrive}, they treat world modeling as an auxiliary semantic task while neglecting generative-level world evolution. This fragmented world foresight (\cref{fig:ECCV_intro}(c)) overlooks the reciprocal interplay between the ego and surrounding agents, failing to synthesize joint interactive trajectories from a world-generative perspective.
(3) \emph{Absence of strategic social reasoning.} Current reasoning mechanisms mainly focus on static scene descriptions \cite{chi2025impromptu,luo2025adathinkdrive}, lacking the game-theoretic `if-what' imagination required for proactive social interaction among traffic participants.
Motivated by these limitations, we seek to answer a key question: \emph{\textbf{How can we endow VLA models with comprehensive world cognition across both semantic-level forecasting and generative-level evolution, thereby enabling proactive driving?}}

To address these challenges, we propose \textbf{WCog-VLA}, a novel VLA framework featuring dual-level \textbf{W}orld \textbf{Cog}nition (\cref{fig:ECCV_intro}(d)). WCog-VLA achieves comprehensive world cognition by bridging the gap between semantic-level forecasting and generative-level physical evolution. 
At the \textit{semantic level}, WCog-VLA first embeds 3D spatial priors into the VLM, establishing a cognitive foundation for structured environment understanding. To operationalize the semantic world understanding, we decouple the VLM's output hidden states into two functional roles. The cognition role encapsulates the understanding of current world states and future world dynamics. Meanwhile, the reasoning role drives a four-step game-theoretic reasoning process. This strategic reasoning mechanism transforms the ego vehicle from a passive observer to a proactive negotiator in social driving scenarios.
At the \textit{generative level}, we introduce the Aligned Decoupled Diffusion Transformer (ADDT) as our generative world model. Conditioned on the VLM's hidden states, ADDT employs a decoupled encoder-decoder architecture to generate multi-agent trajectories. Specifically, the condition encoder explicitly aligns its latent space with dynamic scene representation. Guided by this alignment, the generation decoder efficiently synthesizes joint multi-agent trajectories, firmly grounding ego-planning within predicted interactive future dynamics.
In summary, our main contributions are as follows:

1. \textbf{Dual-level world cognition framework}. We propose WCog-VLA that bridges semantic-level forecasting with generative-level evolution, enabling proactive autonomous driving.

2. \textbf{Generative world model}. We introduce ADDT as a generative world model, which synthesizes physically-plausible joint multi-agent trajectories with efficient inference time.

3. \textbf{Game-theoretic reasoning dataset}. We construct Game-CoT, a Game-theoretic Chain-of-Thought reasoning dataset with 85k samples that fills the gap in game-theoretic reasoning supervision for social driving.

4. \textbf{State-of-the-art (SOTA) performance}. WCog-VLA achieves a SOTA PDMS score of 92.9 on the NAVSIM \cite{dauner2024navsim} benchmark.

\section{Related Work}
\subsubsection{End-to-End Autonomous Driving.}
E2E autonomous driving directly maps sensory inputs to trajectories within unified frameworks \cite{chitta2022transfuser,hu2022st,li2024hydra}. Pioneering works like UniAD \cite{hu2023planning} and VAD \cite{jiang2023vad} integrate perception and planning using dense Bird's-Eye-View (BEV) representations, while SparseAD \cite{sun2025sparsedrive} improves efficiency via sparse queries. Recent paradigms, \eg, GenAD \cite{zheng2024genad}, DiffusionDrive \cite{liao2025diffusiondrive}, and VADv2 \cite{chen2024vadv2}, introduce generative and probabilistic models for multi-modal trajectory planning \cite{zheng2025diffusion}. However, current E2E models are limited by training data coverage, frequently failing in long-tail scenarios. This fragility primarily stems from an inherent lack of semantic reasoning and environment understanding capabilities, motivating the integration of VLMs into E2E driving.

\vspace{-1em}
\subsubsection{VLA for Autonomous Driving.}
VLMs are initially applied as high-level semantic interpreters for scenario understanding, such as DriveGPT4 \cite{xu2024drivegpt4} and DriveLM \cite{tian2024drivevlm}. Building upon this, unified VLA models have been proposed to directly map multi-modal inputs to driving actions \cite{jiang2024senna}, exemplified by EMMA \cite{hwang2024emma,xing2025openemma}, SimLingo \cite{renz2025simlingo}, and DriveMoE \cite{yang2025drivemoe}. While early VLA models driving actions directly as text \cite{liu2025omnireason,luo2025adathinkdrive}, later work such as AutoVLA \cite{zhou2025autovla} introduces autoregressive generation of action tokens. More recently, studies integrate VLM with generative planners \cite{wang2025alpamayo,li2025discrete} to mitigate modal collapse between text and actions, \eg, VAE-based ORION \cite{fu2025orion} and diffusion-based ReCogDrive \cite{li2025recogdrive}.
However, current VLA methods predominantly remain reactive observers. Lacking explicit world cognition and future forecasts, they fail to anticipate dynamic changes in complex social scenarios, motivating the integration of comprehensive world cognition into the VLA framework.

\vspace{-1em}
\subsubsection{World Cognition Building for VLA-based Autonomous Driving.}
The latest VLA studies \cite{liu2026driveworld, xiong2026unidrive,jiang2025irl,xiao2025world} have sought to incorporate world cognition. One branch of research focuses on enhancing spatial awareness of VLMs \cite{dang2026sparseoccvla,han2025percept}, \eg, DrivePI \cite{liu2025drivepi} introduces spatial-aware world cognition, and SGDrive \cite{li2026sgdrive} builds world features around a scene-agent-goal hierarchy. Another branch leverages generative forecasting for future foresight, employing future image generation as an auxiliary objective, \eg, UniDrive-WM \cite{xiong2026unidrive} and DriveVLA-W0 \cite{li2025drivevla}. However, these methods exhibit fragmented world foresight.
Whether relying on semantic or image forecasting, they treat world evolution as a supervised perceptual task and lack joint multi-agent planning, failing to model interactive physical behaviors from a generative world perspective. To bridge this gap, WCog-VLA introduces a unified, dual-level world cognition framework.
At the semantic level, the explicit agent tokens encapsulate the cognition of world dynamics. At the generative level, our ADDT synthesizes joint multi-agent trajectories, manifesting explicit world cognition within the generative execution stage.

\begin{figure}[tb]
  \centering
  \includegraphics[trim={1cm 1.15cm 1cm 1cm}, clip, width=0.95\textwidth]{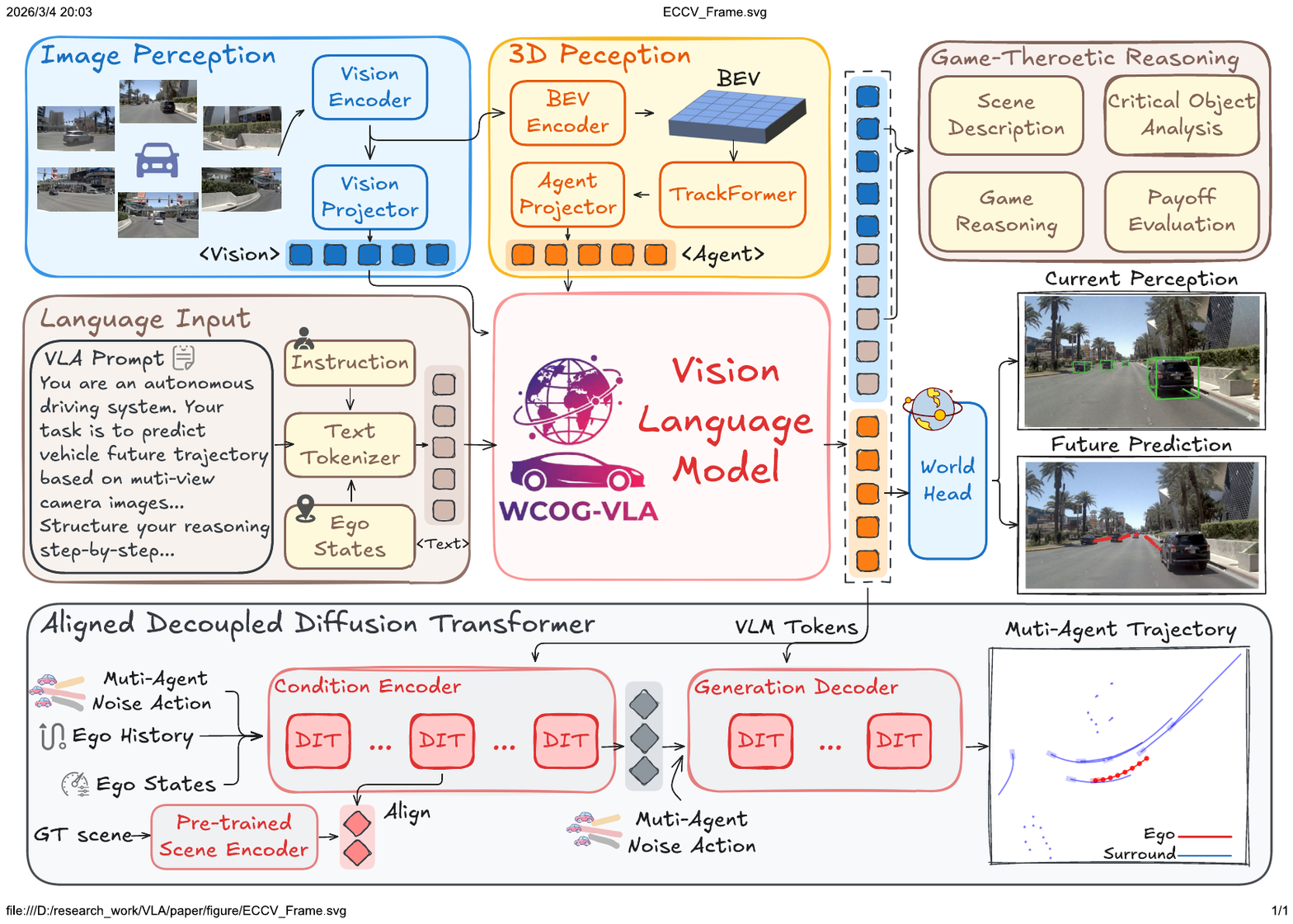}
  \caption{Overview of WCog-VLA. Our framework achieves dual-level world cognition by tightly coupling a multi-modal VLM backbone with a generative world model ADDT. The VLM integrates vision, text, and agent tokens to perform Game-CoT reasoning and semantic world forecasting, while the ADDT translates these cognitive representations to generate physically-plausible, joint multi-agent trajectories.}
  \label{fig:Frame}
  \vspace{-0.75em}
\end{figure}

\section{WCog-VLA}
As shown in \cref{fig:Frame}, our WCog-VLA consists of two tightly coupled components that bridge semantic-world understanding and reasoning with generative trajectory synthesis. First, a \textit{VLM Backbone} is constructed on a multi-modal architecture to jointly process multi-view camera inputs and textual instructions. These heterogeneous tokens are fused to facilitate a game-theoretic Game-CoT reasoning process, enabling structured reasoning over scene context and agent interactions. To explicitly model spatial structure and agent dynamics, the VLM backbone further incorporates agent tokens from a 3D perception module. A specialized world head then decodes the aggregated agent tokens to current 3D perception and future trajectory predictions of surrounding agents, enabling explicit semantic-level world cognition. Second, to seamlessly bridge semantic intent with physically feasible motion, we employ the \textit{ADDT} as a generative world model. By intrinsically aligning its latent space with implicit scene dynamics through a structured conditioning mechanism, ADDT generates high-fidelity, joint multi-agent trajectories. Collectively, the proposed framework consistently translates explicit semantic world cognition into coherent and dynamically plausible trajectory generation. Detailed model architectures are provided in this section.

\subsection{VLM Backbone}
\subsubsection{Model Inputs and Base VLM Model.} The VLM takes multi-view camera images, instructions, and ego-vehicle states as inputs. The visual input $\mathcal{I} = \{I^i\}_{i=1}^6$ comprises six surround-view images. The instruction $l_{\text{ins}}$ provides navigation commands (\eg, `turn right'). The ego state $\mathcal{S} = \{v, a, \mathcal{T}_{\text{hist}}\}$ encapsulates the current velocity $v$, acceleration $a$, and a 2-second historical trajectory $\mathcal{T}_{\text{hist}}$ sampled at 2 Hz. To process these multi-modal inputs, we adopt InternVL3-2B \cite{zhu2025internvl3} as our VLM backbone, utilizing a 300M-parameter InternViT vision encoder and a Qwen2.5 Large Language Model (LLM).

\vspace{-1em}
\subsubsection{3D Spatial Perception.}
We extend 2D VLM perception into the 3D domain to enable explicit 3D spatial perception. Specifically, the multi-view camera features extracted by the vision encoder are lifted into a BEV representation $\mathcal{F}_{\text{BEV}}$ via an off-the-shelf BEV encoder from BEVFormer \cite{li2024bevformer}. 
To extract structured object representations, we employ a TrackFormer \cite{hu2023planning} that maps dense BEV features to sparse agent-centric tokens. By performing cross-attention between learnable agent queries $Q_{\text{agent}}$ and $\mathcal{F}_{\text{BEV}}$, the module extracts a set of $N_a$ agent tokens $\mathcal{T}_{\text{agent}} = \{t_{\text{agent}}^j\}_{j=1}^{N_a}$, where $N_a$ denotes the number of detected agents. These explicit agent tokens capture spatial locations and geometric features, providing structured inputs for subsequent multi-modal reasoning.

\vspace{-1em}
\subsubsection{Unified World Cognition and Reasoning.}
To achieve comprehensive scene understanding, the vision ($\mathcal{T}_{\text{vision}}$), text ($\mathcal{T}_{\text{text}}$), and agent tokens ($\mathcal{T}_{\text{agent}}$) are concatenated along the sequence dimension and fed into the LLM to model multi-modal interaction and fusion. The resulting hidden states are defined as:

\begin{equation}
    O_{\text{vision}}, O_{\text{text}}, O_{\text{agent}} = \text{LLM}([\mathcal{T}_{\text{vision}}, \mathcal{T}_{\text{text}}, \mathcal{T}_{\text{agent}}])
    \label{eq:llm_output}
\end{equation}

We decouple these output tokens of hidden states to support two distinct downstream tasks. Specifically, $O_{\text{agent}}$ encapsulates semantic-level world cognition, which is routed to a specialized world head for current 3D perception and future trajectory prediction of surrounding agents. Meanwhile, $O_{\text{vision}}$ and $O_{\text{text}}$ are processed by the language modeling head to generate textual responses. Trained via our Game-CoT reasoning paradigm, the model can output explicit game-theoretic reasoning processes in its textual responses.

\begin{figure}[tb]
  \centering
  \includegraphics[trim={1cm 13.6cm 1cm 1cm}, clip, width=\textwidth]{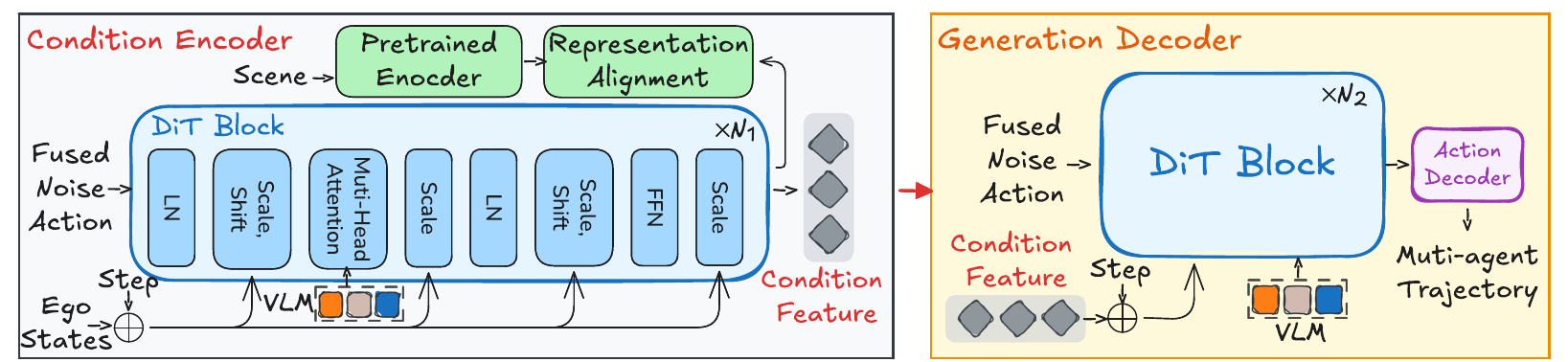}
  \caption{Illustration of the ADDT. The ADDT features a decoupled architecture comprising a condition encoder with representation alignment and a generation decoder.}
  \label{fig:ADDT}
  \vspace{-0.75em}
\end{figure}

\subsection{Aligned Decoupled Diffusion Transformer}
While diffusion transformers exhibit remarkable generation quality, they suffer from an optimization dilemma in single-network architectures \cite{peebles2023scalable}: the encoding of low-frequency abstract semantics conflicts with the decoding of high-frequency continuous details \cite{wang2025ddt}. In autonomous driving, this manifests as a tension between modeling complex multi-agent interactions and generating precise trajectories. To resolve this challenge and bridge VLM semantic cognition with physical actions, we propose ADDT (\cref{fig:ADDT}), which features a decoupled architecture comprising a specialized condition encoder and a dedicated generation decoder.

\vspace{-1em}
\subsubsection{Condition Encoder.} The condition encoder focuses on extracting structural and interactive semantics, decoupled from the burden of precise trajectory recovery. Let $x_{t} \in \mathbb{R}^{N_m \times H \times 3}$ denote the joint multi-agent action noises at diffusion timestep $t$, where $N_m$ is the maximum number of agents and $H$ is the planning horizon. 
To inject temporal and cognitive context, we construct a fused noises representation $F_{at}$ by concatenating embedded noise actions, historical ego actions $\tau_{\text{his}}$, and average-pooled VLM output tokens $\bar{F}_{\text{VLM}}$.
As shown in \cref{fig:ADDT}, $F_{at}$ serves as the primary input for $N_1$ Diffusion Transformer (DiT) blocks. Within these blocks, the diffusion timestep $t$ and ego states $S$ are injected via AdaLN modulation to provide physical kinematics guidance. Concurrently, the full-sequence VLM output tokens $F_{\text{VLM}}=[O_{\text{vision}}, O_{\text{text}}, O_{\text{agent}}]$ are injected into cross-attention layers, providing high-level semantic cognition priors. Finally, the encoder outputs a semantic self-condition feature $z_t$, formulated as:
\vspace{-0.25em}
\begin{equation}
z_t = \text{Encoder}(F_{at}, t, S, F_{\text{VLM}}),\text{  }F_{at} = \text{concat}(E_{act}(x_t), E_{his}(\tau_{\text{his}}), \bar{F}_{\text{VLM}})
\label{eq:encoder}
\end{equation}

The resulting feature $z_t$ captures semantic scene dynamics, which guides the subsequent trajectory generation process.

\vspace{-1em}
\subsubsection{Representation Alignment.} To ensure $z_t$ adheres strictly to real-world dynamics, we introduce a representation alignment mechanism. Specifically, the intermediate feature $h_i$ from the $i$-th DiT block of the condition encoder is aligned with a latent scene representation $r_*$ extracted from a pre-trained VAE encoder, thereby reducing the `semantic' gap between condition encoder output and latent scene space. Following GenAD \cite{zheng2024genad}, this VAE is pre-trained to reconstruct multi-agent trajectories via an MLP encoder and GRU decoder. The resulting latent space after VAE encoder captures both global traffic patterns and individual characteristics of each agent. We enforce this alignment using a cosine similarity constraint \cite{yu2024representation} with a learnable projection MLP $h_\phi$:
\begin{equation}
\mathcal{L}_{\text{align}} = 1 - \cos(r_*, h_\phi(h_i))
\label{eq:align}
\end{equation}

Crucially, this explicit alignment acts as a regularization technique, maintaining the local consistency of $z_t$ across adjacent denoising timesteps. It ensures that the generated trajectories are grounded in feasible scene dynamics while stabilizes semantic features throughout the progressive denoising steps.

\vspace{-1em}
\subsubsection{Generation Decoder.} The generation decoder, comprising $N_2$ DiT blocks, shares the condition encoder's architecture but focuses exclusively on recovering high-frequency geometric details. Guided by the self-condition feature in $z_t$, it processes the fused action noises $F_{at}$ and VLM output tokens $F_{\text{VLM}}$ to estimate the denoised multi-agent trajectories. Unlike the condition encoder, the generation decoder injects both the timestep $t$ and the self-condition feature $z_t$ via AdaLN modulation, enabling semantically aligned denoising. The decoding process is formulated as:

\begin{equation}
x_{t-1} = \text{Decoder}(F_{at}, t, z_t, F_{\text{VLM}})
\label{eq:decoder}
\end{equation}

\vspace{-0.2em}
\subsection{Game-CoT Reasoning Annotation}
Existing reasoning datasets often lack social interaction logic and game-theoretic analysis. To bridge this gap, we propose an automated annotation pipeline powered by advanced Qwen3-VL-Plus to generate structured reasoning across four sequential steps: (1) scene description, (2) critical object analysis, (3) game-theoretic reasoning, and (4) payoff evaluation. The final output includes the optimal ego action and the inferred responses of surrounding agents. Specifically, the game-theoretic reasoning step formulates traffic interactions as a Stackelberg game \cite{stackelberg}, where the ego vehicle acts as the leader and surrounding agents serve as followers. Adopting a `if-what' imagination, the model enumerates candidate ego actions and infers the corresponding reactions of followers. The payoff evaluation step then assesses the safety and efficiency of these hypothetical outcomes to determine the optimal strategy.

To minimize hallucinatory outputs and ensure logical consistency, we incorporate Ground-Truth (GT) actions as guiding hints. This compels the VLM to reconstruct explicit causal chains linking observed scene contexts to final GT actions. Ultimately, we construct a Game-CoT dataset comprising 85k high-quality annotations on the NAVSIM benchmark. More detailed Game-CoT annotation process is illustrated in the Supplemental Material.

\subsection{WCog-VLA Training}
\begin{figure}[tb]
  \centering
  \includegraphics[trim={1cm 4.1cm 1cm 1cm}, clip, width=0.90\textwidth]{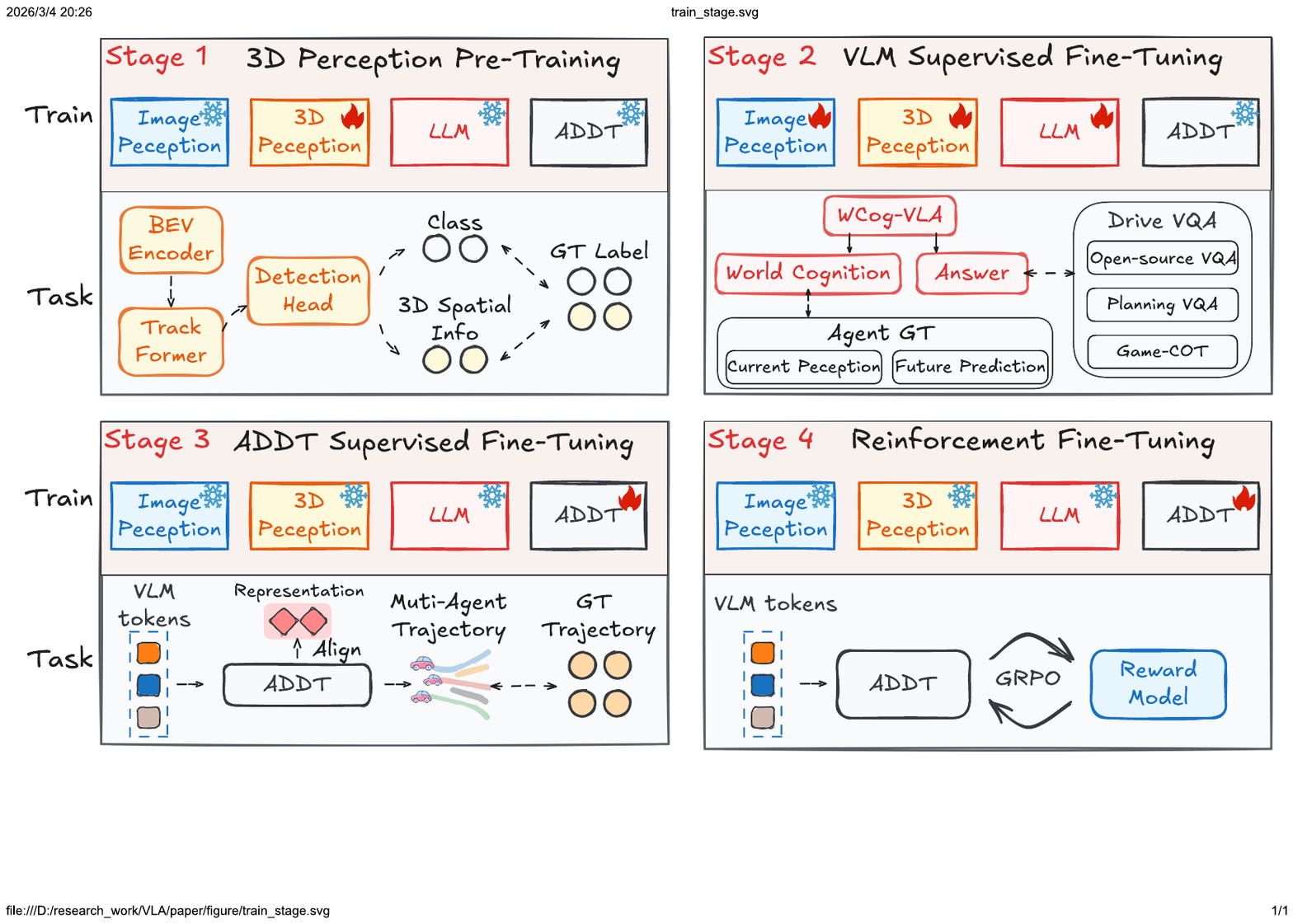}
  \caption{Illustration of four-stage training paradigm of WCog-VLA, including three-stage supervised fine-tuning and one-stage reinforcement fine-tuning.}
  \label{fig:train_stage}
  \vspace{-0.75em}
\end{figure}

\cref{fig:train_stage} shows our four-stage training paradigm, including Supervised Fine-Tuning (SFT) in Stages 1–3 and Reinforcement Fine-Tuning (RFT) in Stage 4. 

\vspace{-1em}
\subsubsection{3D Perception Pre-Training.} This stage optimizes the BEV encoder and TrackFormer. Following the perception training method in UniAD \cite{hu2023planning}, we utilize a detection head for class labeling and 3D box regression. The overall detection loss combines a focal loss for classification and an $L_1$ loss for 3D box localization: 
$\mathcal{L}_{\text{s1}} = \lambda_{\text{focal}} \mathcal{L}_{\text{focal}} + \lambda_{\text{L1}} \mathcal{L}_{\text{L1}}$ with  weighting coefficients $\lambda_{\text{focal}}$ and $\lambda_{\text{L1}}$.

\vspace{-1em}
\subsubsection{VLM Supervised Fine-Tuning.}
This stage optimizes the VLM for both Visual Question Answering (VQA) capability and world cognition. The language modeling is trained on a mixture of public driving VQA datasets (\eg, DriveLM), trajectory-specific VQA, and our Game-CoT dataset.
The learned semantic world cognition are routed through a world head into current perception and future prediction of surrounding agents, which are supervised by ground-truth 3D bounding boxes, and future trajectories.
The overall training objective combines the text generation loss $\mathcal{L}_{\text{LM}}$ and the world cognition loss $\mathcal{L}_{\text{world}}$:
\begin{equation}
\mathcal{L}_{\text{s2}} = \mathcal{L}_{\text{LM}} + \lambda_{\text{world}} \mathcal{L}_{\text{world}}
\label{eq:stage2_loss}
\end{equation}

\vspace{-1em} 
\begin{equation}
\mathcal{L}_{\text{world}} = \frac{1}{N_a} \sum_{i=1}^{N_a} \lambda_{\text{box}} \mathcal{L}_{\text{L1}}(b_i) + \frac{1}{N_a H} \sum_{i=1}^{N_a} \sum_{t=1}^{H} \lambda_{\text{traj}} \mathcal{L}_{\text{L1}}(\tau_{i,t})
\label{eq:loss_world}
\end{equation}
where $\mathcal{L}_{\text{LM}}$ is the standard cross-entropy loss for language modeling, $\mathcal{L}_{\text{L1}}(b_i)$ and $\mathcal{L}_{\text{L1}}(\tau_{i,t})$ denote the $L_1$ losses for 3D box localization and trajectory prediction of agent $i$ at timestep $t$. The $\lambda$ terms denote the corresponding weights.

\vspace{-1em}
\subsubsection{ADDT Supervised Fine-Tuning.}
 In Stage 3, we freeze the pre-trained VLM to serve as a semantic-level world model and train the ADDT for joint multi-agent trajectory generation. The training objective combines the standard $L_2$ denoising loss $\mathcal{L}_{\text{diff}}$ and the representation alignment loss $\mathcal{L}_{\text{align}}$:
\begin{equation} 
    \mathcal{L}_{\text{s3}} = \mathcal{L}_{\text{diff}} + \lambda_{\text{align}} \mathcal{L}_{\text{align}} 
    \label{eq:loss_gil} 
\end{equation}

\vspace{-0.5em} 
\begin{equation} 
    \mathcal{L}_{\text{diff}} = \mathbb{E}_{z_{t}, F_{at}, \epsilon \sim \mathcal{N}(0,I)} \left[ \| \mathbf{W} \odot \big( \epsilon - \epsilon_\theta(F_{at}, t, z_t, F_{\text{VLM}}) \big) \|_2^2 \right] 
    \label{eq:loss_diff} 
\end{equation}
where $\mathcal{L}_{\text{diff}}$ optimizes the generation decoder $\epsilon_\theta$ to predict the added noise $\epsilon$, and $\odot$ is the Hadamard product. $\mathbf{W}$ is an agent-specific weight mask applying distinct penalties ($\alpha_{\text{ego}}$ and $\alpha_{\text{surr}}$) to prioritize the generative accuracy of the ego vehicle over surrounding agents and $\lambda_{\text{align}}$ is the weighting coefficient.

\vspace{-1em}
\subsubsection{Reinforcement Fine-Tuning.} 
To enable driving exploration beyond imitation \cite{huang2025vlm}, we adopt DiffGRPO \cite{li2025recogdrive,guo2025deepseek}, a diffusion-specific GRPO algorithm. The DiffGRPO loss includes an RL policy optimization loss and a Behavior Cloning (BC) loss to prevent policy collapse during exploration:
\begin{equation}
\resizebox{0.93\linewidth}{!}{$
    L_{\text{s4}} = \underbrace{-\frac{1}{GT} \sum_{i=1}^{G} \sum_{t=1}^{T} \gamma^{t-1} \log \pi_\theta (x_{t-1}^{(i)} \mid x_t^{(i)}) \hat{A}_i}_{L_{\text{RL}}} - \underbrace{\lambda_{bc} \frac{1}{GT} \sum_{i=1}^{G} \sum_{t=1}^{T} \log \pi_\theta (\tilde{x}_{t-1}^{(i)} \mid \tilde{x}_t^{(i)})}_{L_{\text{BC}}}
$}
\label{eq:rl_loss}
\end{equation}
where $G$ and $T$ denote the sampled group size and total denoising steps, respectively. $\gamma$ is the discount coefficient mitigating instability in early denoising steps, $\pi_\theta(x_{t-1}^{(i)} \mid x_t^{(i)})$ is each step's conditional probability. $\hat{A}_i$ is the group-relative advantage, $\lambda_{bc}$ is the weight of the BC loss. $\tilde{x}_{t-1}^{(i)}$ and $\tilde{x}_t^{(i)}$ are sampled from the reference policy $\pi_{\text{ref}}$ (\eg, the model after SFT).

We design a joint reward function decoupling the ego vehicle and surrounding agents. Ego driving quality is evaluated via the NAVSIM Predictive Driving Model Score (PDMS), whereas surrounding agents are optimized for accurate motion forecasting via a negative $L_1$ displacement penalty.  The overall reward is formulated as $r_i = r_{\text{PDMS}} - \lambda_{\text{surr}} \mathcal{L}_{\text{L1}}(\tau_{\text{surr}})$ where $\lambda_{\text{surr}}$ balances ego planning performance with consistent surrounding agents' trajectory prediction.

\section{Experiments}

\subsection{Experimental setup}
\subsubsection{Dataset.} We evaluate WCog-VLA on the large-scale, real-world simulation benchmarks NAVSIMv1 and NAVSIMv2 \cite{dauner2024navsim}. NAVSIM is a planning-oriented dataset comprising challenging scenarios. The data is partitioned into 1,192 training (\textit{navtrain}) and 136 test (\textit{navtest}) scenes. To establish the foundational driving cognition of the VLM, we compile a comprehensive training mixture comprising over 158k samples from open-source driving VQA datasets, including DriveLM \cite{tian2024drivevlm}, CODA-LM \cite{chen2025coda-lm}, LingoQA \cite{marcu2024lingoqa}, nuScenes-QA \cite{qian2024nuscenes}, NuInstruct \cite{ding2024NuInstruc}, and DriveGPT4 \cite{xu2024drivegpt4}. This corpus is further augmented with 170K NAVSIM-tailored samples, including 85k trajectory-specific VQA and 85k Game-CoT reasoning samples.

\vspace{-1em}
\subsubsection{Implementation Details.}
Our training pipeline consists of four sequential stages. First, the 3D perception module is trained on NAVSIM for 1 epoch. Second, the VLM is pre-trained for 1 epoch on the 158k VQA samples, followed by 3 epochs of joint fine-tuning with the world heads using the 170K NAVSIM-tailored samples. Third, with the VLM frozen, ADDT is trained via DDPM \cite{ho2020denoising} for 200 epochs on NAVSIM. The ADDT features a symmetric 16-block DiT architecture (8 blocks each for the condition encoder and generation decoder). Representation alignment is extracted from the 6th DiT encoder block. Finally, ADDT is refined via GRPO on NAVSIM for 10 epochs using 6 group sizes. All training stages are conducted on 4 NVIDIA A100 40GB GPUs. Additional implementation details are provided in the Supplemental Material.

\subsection{Main Results}
\subsubsection{Results on NAVSIM v1.}
\cref{tab:NAVSIM_v1_result} presents the closed-loop evaluation of our WCog-VLA on the NAVSIM v1. Our method achieves a SOTA PDMS of 92.9, outperforming all listed standard end-to-end and VLM-based methods. Notably, despite using only camera inputs, WCog-VLA surpasses multi-modal baselines that leverage both camera and lidar inputs, yielding an improvement of 4.6 PDMS over WoTE. Furthermore, WCog-VLA demonstrates clear advantages over VLM-based methods. It outperforms two massive generalist models, QwenVL2.5 and InternVL3, by 9.6 PDMS, validating the effectiveness of our tailored architecture and driving knowledge injection. Crucially, our compact 2B model surpasses the RL-refined VLM-based methods, ReCogDrive and AutoVLA, by at least 0.8 PDMS and outperforms the 3B-parameter LatentVLA by 0.5 PDMS. These results underscore that enhancing VLA with world cognition and world-dynamics forecasting enables superior planning performance. 

Beyond overall planning performance, WCog-VLA excels in safety metrics, achieving a remarkable 99.4 in NC and 98.5 in TTC. This is because our model anticipates the future intents of surrounding agents, enabling proactive safety measures and collision avoidance in complex scenarios.

\begin{table}[tb]
    \caption{Performance comparison on NAVSIM v1 \emph{navtest}. Our WCog-VLA is evaluated after the complete four-stage training. Metrics include NC (no at-fault collision), DAC (drivable area compliance), TTC (time-to-collision), Comf. (comfort), EP (ego progress), and PDMS (predictive driver model score). $\dagger$ indicates models fine-tuned on the NAVSIM trajectory-specific dataset.}
    \label{tab:NAVSIM_v1_result}
    \small 
    \renewcommand{\arraystretch}{0.85}
    \setlength{\tabcolsep}{4pt} 
    \resizebox{\textwidth}{!}{
    \begin{tabular}{l|cc|cc|ccc|>{\columncolor{tablegray}}c}
        \toprule
        Method & Image & Lidar & NC$\uparrow$ & DAC$\uparrow$ & TTC$\uparrow$ & Comf. $\uparrow$ & EP$\uparrow$ & \cellcolor{white}PDMS$\uparrow$ \\ 
        \midrule
        Constant Velocity & & & 68.0 & 57.8 & 50.0 & \textbf{100} & 19.4 & 20.6 \\
        Ego Status MLP    & & & 93.0 & 77.3 & 83.6 & \textbf{100} & 62.8 & 65.6 \\
        \midrule
        VADv2-$\mathcal{V}_{8192}$ \cite{chen2024vadv2} & $\checkmark$ & & 97.2 & 89.1 & 91.6 & \textbf{100} & 76.0 & 80.9 \\
        DrivingGPT \cite{chen2025drivinggpt} & $\checkmark$ & & 98.9 & 90.7 & 94.9 & 95.6 & 79.7 & 82.4 \\
        UniAD \cite{hu2023planning} & $\checkmark$ & & 97.8 & 91.9 & 92.9 & \textbf{100} & 78.8 & 83.4 \\
        BevDrive \cite{yu2025combining}& $\checkmark$ & $\checkmark$ & 97.7 & 92.5 & 92.9& \textbf{100} & 78.7 & 83.8 \\
        TransFuser \cite{chitta2022transfuser}& $\checkmark$ & $\checkmark$ & 97.7 & 92.8 & 92.8 & \textbf{100} & 79.2 & 84.0 \\
        PARA-Drive \cite{weng2024drive}& $\checkmark$ & & 97.9 & 92.4 & 93.0 & 99.8 & 79.3 & 84.0 \\
        DRAMA \cite{yuan2024drama}& $\checkmark$ & $\checkmark$ & 98.0 & 93.1 & 94.8 & \textbf{100} & 80.1 & 85.5 \\
        Hydra-MDP-$\mathcal{V}_{8192}$-W-EP \cite{li2024hydra}& $\checkmark$ & $\checkmark$ & 98.3 & 96.0 & 94.6 & \textbf{100} & 78.7 & 86.5 \\
        DiffusionDrive \cite{liao2025diffusiondrive}& $\checkmark$ & $\checkmark$ & 98.2 & 96.2 & 94.7 & \textbf{100} & 82.2 & 88.1 \\
        WoTE \cite{li2025end}& $\checkmark$ & $\checkmark$ & 98.5 & 96.8 & 94.9 & 99.9 & 81.9 & 88.3 \\
        iPad \cite{guo2025ipad}& $\checkmark$ & & 98.6 & 98.3 & 94.9 & \textbf{100} & 88.0 & 91.7 \\
        \midrule
        \multicolumn{9}{l}{\textbf{VLMs-based Methods}} \\
        QwenVL2.5-8B \cite{bai2025qwen3}$^\dagger$ & $\checkmark$ & & 97.8 & 92.1 & 92.8 & \textbf{100} & 78.3 & 83.3 \\
        InternVL3-8B \cite{zhu2025internvl3}$^\dagger$ & $\checkmark$ & & 97.0 & 92.4 & 91.8 & \textbf{100} & 78.9 & 83.3 \\
        ReCogDrive-2B \cite{li2025recogdrive}& $\checkmark$ & & 97.9 & 97.3 & 94.9 & \textbf{100} & 87.3 & 90.8 \\
        AutoVLA-3B \cite{zhou2025autovla}& $\checkmark$ & & 99.1 & 97.1 & 97.1 & \textbf{100} & 87.6 & 92.1 \\
        LatentVLA-3B \cite{xie2026latentvla}& $\checkmark$ & & 98.9 & 98.2 & 95.2 & \textbf{100} & \textbf{88.2} & 92.4 \\
        \midrule
        WCog-VLA-2B(ours) & $\checkmark$ & & \textbf{99.4} & \textbf{98.8} & \textbf{98.5} & \textbf{100} & 87.1 & \textbf{92.9} \\
        \bottomrule
    \end{tabular}
    }

\end{table}

\vspace{-1em}
\subsubsection{Results on NAVSIM v2.}
\cref{tab:NAVSIM_v2_result} presents the evaluation on the NAVSIM v2 benchmark, with WCog-VLA deployed after three-stage SFT process. Our WCog-VLA achieves a SOTA Extended PDMS (EPDMS) of 85.9, outperforming DiffusionDrive by 1.6 EPDMS. Besides, WCog-VLA attains the highest safety scores in both NC and TTC, while maintaining highly competitive performance across all other metrics. These findings further confirm the effectiveness and robust generalization capability of WCog-VLA in extended driving evaluations.

\begin{table}[tb]
    \centering
    \caption{Performance comparison on NAVSIM v2 \emph{navtest} with extended metrics. Our WCog-VLA is evaluated after three-stage SFT. Newly introduced metrics include DDC (driving direction compliance), TLC (traffic light compliance), LK (lane keeping), HC (history comfort), EC (extended comfort), and EPDMS (extended PDMS).}
    \label{tab:NAVSIM_v2_result}

    \setlength{\tabcolsep}{3pt}
    \small 
    \renewcommand{\arraystretch}{0.90}
    \resizebox{\textwidth}{!}{
        \begin{tabular}{l|cccc|ccccc|>{\columncolor{tablegray}}c}
            \toprule
            Method & NC $\uparrow$ & DAC $\uparrow$ & DDC $\uparrow$ & TLC $\uparrow$ & EP $\uparrow$ & TTC $\uparrow$ & LK $\uparrow$ & HC $\uparrow$ & EC $\uparrow$ & \multicolumn{1}{c}{EPDMS $\uparrow$} \\ 
            \midrule
            VADv2 \cite{chen2024vadv2}& 97.3 & 91.7& 98.2 & 99.9 & 77.6 & 92.7 & 66.0& \textbf{100} & 97.4 & 76.6 \\
            TransFuser \cite{chitta2022transfuser}& 97.7 & 92.8 & 98.3 & 99.9 & 79.2 & 92.8 & 67.6& \textbf{100} & 95.3 & 77.8 \\
            HydraMDP++ \cite{li2024hydra}& 97.9 & 96.5 & 98.9 & \textbf{100} & 79.2 & 93.4 & 67.2 & \textbf{100} & 97.7 & 80.6 \\
            ARTEMIS \cite{feng2025artemis}& 98.3 & 95.1 & 98.6 & 99.8 & 81.5 & 97.4 & 96.5 & \textbf{100} & \textbf{98.3} & 83.1 \\
            ReCogDrive-8B \cite{li2025recogdrive}& 98.3 & 95.2 & \textbf{99.5} & 99.8 & 87.1 & 97.5 & 96.6 & 98.3 & 86.5 & 83.6 \\
            WoTE \cite{li2025end}& 98.5 & \textbf{96.8} & 98.8 & 99.8 & 86.1 & 97.9 & 95.5 & 98.3 & 82.9 & 84.2 \\
            DiffusionDrive \cite{liao2025diffusiondrive}& 98.0 & 96.0 & \textbf{99.5} & 99.8 & \textbf{87.7} & 97.1 & \textbf{97.2} & 98.3 & 87.6 & 84.3 \\
            \midrule
            WCog-VLA-2B(ours) &\textbf{98.8}  & 96.6 &  99.3& 99.8 & 85.8 & \textbf{98.2} & 96.4 & 98.3 & 86.3 & \textbf{85.9} \\
            \bottomrule
        \end{tabular}
    }
\end{table}

\subsection{Ablation Study}
\subsubsection{Effect of the Four-Stage Training.}
\cref{tab:ablation_stages} ablates our four-stage training paradigm. Using only Stage 2 yields a baseline PDMS of 84.4, whereas incorporating 3D perception pre-training improves 1.1 PDMS, reflecting enhanced spatial understanding. Introducing ADDT in Stage 3 fundamentally shifts the paradigm from discrete textual output to continuous trajectory generation, resulting in a 3.8 PDMS improvement. Finally, Stage 4 RFT further optimizes the driving policy, improving 3.6 PDMS and achieving the SOTA 92.9. These consistent gains confirm that every training stage is indispensable.

\begin{table*}[tb] %
    \centering
    \footnotesize
    \begin{minipage}[t]{0.45\linewidth}
        \centering
        \caption{Ablation on the four-stage training process. Trajectories are generated as textual tokens via the VLM in IDs 1 and 2, and as continuous actions through ADDT in IDs 3 and 4.}
        \label{tab:ablation_stages}
        \setlength{\tabcolsep}{1.9pt} 
        \renewcommand{\arraystretch}{1.25}
        \resizebox{\textwidth}{!}{
            \begin{tabular}{l | cccc | >{\columncolor{tablegray}}c}
                \toprule
                ID & Stage 1 & Stage 2 & Stage 3 & Stage 4 & \multicolumn{1}{c} {PDMS $\uparrow$} \\
                \midrule
                1 & & $\checkmark$ & & & 84.4 \\
                2 & $\checkmark$ & $\checkmark$ & & & 85.5 \\
                3 & $\checkmark$ & $\checkmark$ & $\checkmark$ & & 89.3 \\
                4 & $\checkmark$ & $\checkmark$ & $\checkmark$ & $\checkmark$ & \textbf{92.9} \\
                \bottomrule
            \end{tabular}
        }
    \end{minipage}
    \hfill 
    \begin{minipage}[t]{0.53\linewidth}
        \centering
        \caption{Ablation on dual-level world cognition. Cur and Fut denote current perception and future prediction supervision. Generative enables joint multi-agent trajectory synthesis; otherwise, only ego-trajectory generated.}
        \label{tab:ablation_semantic_generative}
        \renewcommand{\arraystretch}{0.55} 
        \setlength{\tabcolsep}{10pt}
        \resizebox{\textwidth}{!}{
            \begin{tabular}{c | cc | c | >{\columncolor{tablegray}}c}
                \toprule
                \multirow{2}{*}{ID} & \multicolumn{2}{c|}{Semantic} & \multirow{2}{*}{Generative} & \multicolumn{1}{c}{\multirow{2}{*}{PDMS $\uparrow$}} \\
                \cmidrule(lr){2-3}
                & Cur & Fut & & \multicolumn{1}{c}{} \\
                \midrule
                1 &              &              &              & 86.5\\
                2 & $\checkmark$ &              &              & 87.0\\
                3 &              & $\checkmark$ &              & 87.2\\
                4 & $\checkmark$ & $\checkmark$ &              & 88.1\\
                5 &              &              & $\checkmark$ & 87.4\\
                6 & $\checkmark$ & $\checkmark$ & $\checkmark$ & \textbf{89.3}\\
                \bottomrule
            \end{tabular}
        }
    \end{minipage}
\vspace{-0.5em}
\end{table*}

\vspace{-1em}
\subsubsection{Effect of Dual-Level World Cognition.}
\cref{tab:ablation_semantic_generative} evaluates the effect of dual-level world cognition, where all variants are trained with the three-stage SFT. The baseline without either cognitive level achieves 86.5 PDMS. Integrating semantic current perception or future prediction improves the score to 87.0 and 87.2, respectively, while combining both yields 88.1. Enabling only generative-level multi-agent synthesis without semantic cognition achieves 87.4. Ultimately, unifying both semantic with generative cognition triggers a synergistic leap to 89.3 PDMS. These findings demonstrate that coupling semantic forecasting with generative evolution is essential for robust planning.

\vspace{-1em}
\subsubsection{Effect of ADDT.}
\cref{tab:ablation_denoise} evaluates our ADDT design in terms of PDMS and inference time. We compare pure VLM text generation, including direct answer ($\text{VLM}^{\text{wo/r}}$) and Game-CoT reasoning ($\text{VLM}^{\text{w/r}}$), against a Standard Diffusion Transformer (SDT) \cite{peebles2023scalable} and several ADDT variants: without alignment (DDT), without the decoupled architecture (ADT), and the full ADDT. All diffusion models utilize an identical DiT backbone and are trained with three-stage SFT. Results show that ADDT with 5 denoising steps achieves a 10.7$\times$ speedup over direct VLM text generation. Crucially, ADDT attains superior performance with fewer denoising steps. Compared to the 20-step SDT, our 5-step ADDT improves PDMS by 0.8 while accelerating inference by 3.7$\times$. Besides, ADDT exhibits low sensitivity to the number of denoising steps: increasing the steps from 5 to 20 yields a marginal 0.3 PDMS gain. This robustness stems from the explicit alignment mechanism, which maintains consistent encoder latent features across different denoising steps and reduces the need for costly iterative refinement.

\begin{table*}[tb]
    \centering
    \begin{minipage}[t]{0.60\linewidth}
        \centering
        \caption{Effect of ADDT. wo/r means VLM text output without reasoning, w/r is with reasoning.}
        \label{tab:ablation_denoise}
        \renewcommand{\arraystretch}{0.58} 
        \setlength{\tabcolsep}{5Pt}
        \resizebox{\linewidth}{!}{
            \begin{tabular}{c | c | c | c}
                \toprule
                Method & Denoise step & PDMS $\uparrow$ & Infer Time (s) $\downarrow$ \\
                \midrule
                $\text{VLM}^{\text{wo/r}}$ & & 85.0 & 1.131 \\
                $\text{VLM}^{\text{w/r}}$ & & 85.5 & 9.896 \\
                \midrule
                \multirow{2}{*}{VLM+SDT\cite{peebles2023scalable}}  & 5  & 87.4 & 0.105 \\
                                      & 20 & 88.5 & 0.388 \\
                \midrule
                \multirow{2}{*}{VLM+DDT}  & 5  & 87.9 & 0.108 \\
                                      & 20 & 88.7 & 0.381 \\
                \midrule
                \multirow{2}{*}{VLM+ADT}  & 5  & 88.6 & \textbf{0.103} \\
                                      & 20 & 89.1 & 0.392 \\
                \midrule
                \multirow{2}{*}{VLM+ADDT} & 5  & 89.3 & 0.106 \\
                                      & 20 & \textbf{89.6} & 0.383 \\
                \bottomrule
            \end{tabular}
        }
    \end{minipage}
    \hfill 
    \begin{minipage}[t]{0.36\linewidth}
        \begin{minipage}[t]{\linewidth}
            \centering
            \small
            \caption{Effect of VQA dataset.}
            \label{tab:ablation_VQA}
            \renewcommand{\arraystretch}{0.80} 
            \setlength{\tabcolsep}{5pt}
            \resizebox{\linewidth}{!}{
                \begin{tabular}{c|ccc|c}
                    \toprule
                    ID & Traj & Drive & CoT & PDMS $\uparrow$ \\
                    \midrule
                    1 & $\checkmark$ &              &              & 86.7 \\
                    2 & $\checkmark$ & $\checkmark$ &              & 88.2 \\
                    3 & $\checkmark$ &              & $\checkmark$ & 87.5 \\
                    4 & $\checkmark$ & $\checkmark$ & $\checkmark$ & \textbf{89.3} \\
                    \bottomrule
                \end{tabular}
            }
        \end{minipage}

        \begin{minipage}[t]{\linewidth}
            \centering
            \footnotesize
            \caption{3D perception effect.}
            \label{tab:ablation_3d_perception}
            \setlength{\tabcolsep}{5pt}
            \renewcommand{\arraystretch}{0.3}
            \resizebox{0.9\linewidth}{!}{ 
                \begin{tabular}{c|c}
                    \toprule
                    3D perception & PDMS $\uparrow$ \\
                    \midrule
                    $\times$ & 86.0 \\
                    $\checkmark$ & \textbf{89.3} \\
                    \bottomrule
                \end{tabular}
            }
        \end{minipage}
    \end{minipage}
\vspace{-0.5em}
\end{table*}

\vspace{-1em}
\subsubsection{Effect of VQA Dataset.}
\cref{tab:ablation_VQA} shows that training solely on NAVSIM trajectory-specific VQA yields a baseline PDMS of 86.7. Adding open-source driving VQA (Drive) or Game-CoT reasoning data (CoT) improves PDMS to 88.2 and 87.5. Combining all three data sources achieves the highest PDMS of 89.3, confirming that incorporating diverse VQA data can improve performance.

\vspace{-1em}
\subsubsection{Effect of 3D Perception.}
\cref{tab:ablation_3d_perception} validates the contribution of the 3D perception module. Without explicit 3D perception, ADDT relies solely on generic VLM vision tokens, which limits spatial precision and yields a PDMS of 86.0. Incorporating the dedicated 3D perception boosts performance to 89.3.

More ablation study results (\eg, effect of alignment layer position) are shown in the \textbf{Supplemental Material}.

\begin{figure}[tb]
  \centering
  \includegraphics[trim={6.5cm 22.9cm 0cm 0cm}, clip, width=\textwidth]{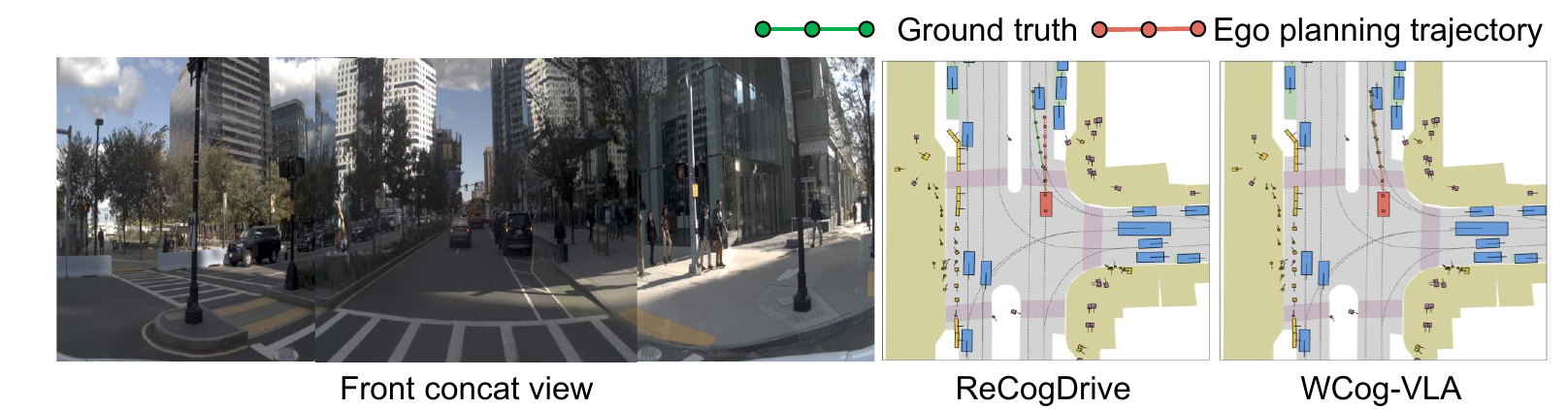}
  \caption{Comparison with previous SOTA method on \emph{Navtest}.}
  \label{fig:qualitative1}
  \vspace{0.5em} 
  \includegraphics[trim={6.5cm 22.5cm 0cm 0cm}, clip, width=\textwidth]{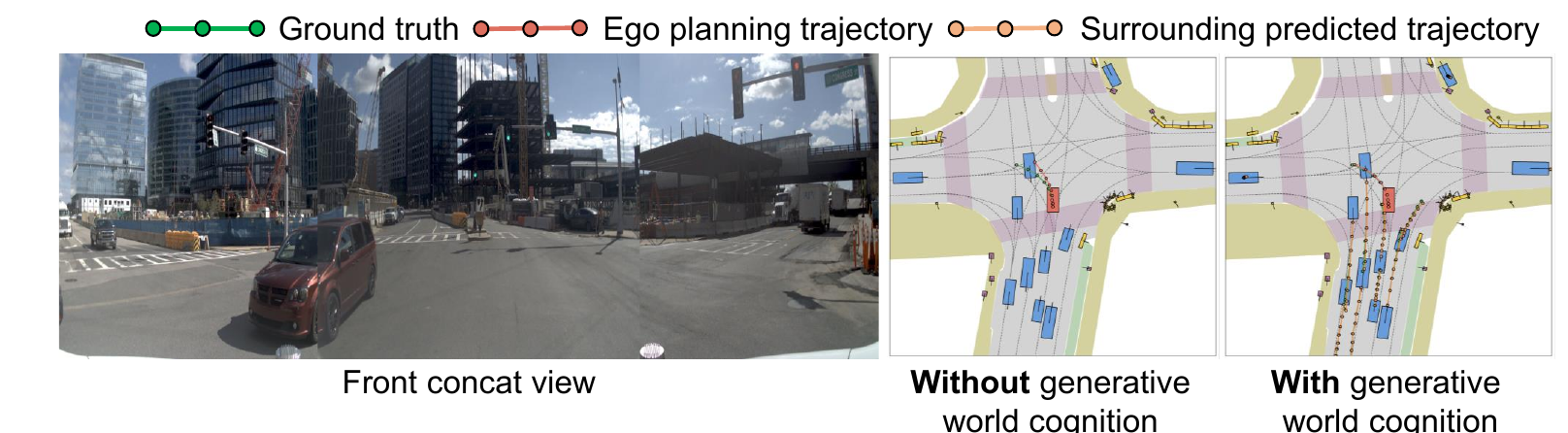}
  \caption{Visualization of proactive driving via generative-level world cognition.}
  \label{fig:qualitative2}
  \vspace{0.5em} 
  \includegraphics[trim={0cm 23.8cm 0cm 0cm}, clip, width=\textwidth]{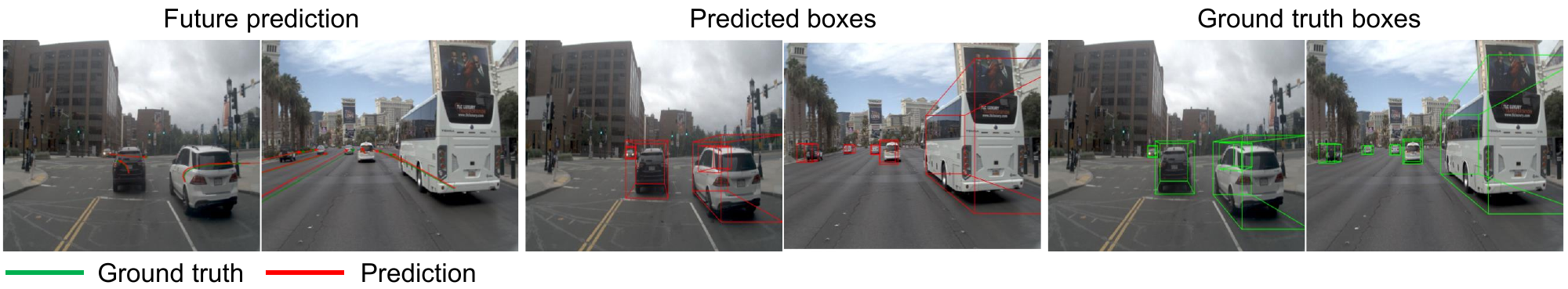}
  \caption{Visualization of our semantic-level world cognition.}
  \label{fig:qualitative3}
  \vspace{-0.75em}
\end{figure}

\subsection{Qualitative Results}
\subsubsection{Compare with Previous SOTA Method.}
\cref{fig:qualitative1} compares WCog-VLA with ReCogDrive \cite{li2025recogdrive} in complex urban scenarios. ReCogDrive acts overly conservatively, remaining trapped in the current slow lane. Conversely, WCog-VLA identifies the slow leading bus in the current lane and changes the lane to improve efficiency, closely matching the human ground truth. This visualization demonstrates our WCog-VLA enables efficient and human-aligned driving behaviors.

\vspace{-1em}
\subsubsection{Proactive Driving via Generative-Level World Cognition.}
\cref{fig:qualitative2} highlights the advantages of generative world cognition. In the intersection scenario, the baseline lacks interactive foresight of the oncoming vehicle and generates an ego-only trajectory, resulting in passive deceleration to blindly avoid spurious conflicts. Conversely, our model synthesizes joint multi-agent trajectories that explicitly forecast the oncoming vehicle's straight trajectory. This foresight enables ego vehicle to confidently execute a left turn. The results confirm that generative world cognition empowers the model to perform proactive maneuvers.

\vspace{-1em}
\subsubsection{Explicit Semantic-Level World Cognition Representation.}
\cref{fig:qualitative3} shows the semantic world cognition decoded via the world head. Both 3D perception and future trajectory prediction align closely with the ground truth, demonstrating the model's cognition of current world states and future world dynamics.

More qualitative results are shown in the \textbf{Supplemental Material}.

\section{Conclusion}
In this work, we propose \textbf{WCog-VLA}, a novel VLA framework with explicit dual-level \textbf{W}orld \textbf{Cog}nition for end-to-end autonomous driving. To bridge the gap between semantic-level forecasting and generative-level evolution, WCog-VLA tightly couples a multi-modal VLM backbone with a generative Aligned Decoupled Diffusion Transformer (ADDT). At the semantic level, our model unifies the world cognition and reasoning, enabling comprehensive world understanding and interactive game-theoretic reasoning. At the generative level, ADDT acts as a generative world model, translates the VLM's cognitive representations into physically-plausible joint multi-agent trajectories. Extensive experiments on the NAVSIM v1 and v2 benchmarks demonstrate SOTA performance of WCog-VLA. 
However, the current semantic cognition focuses on agents and omits the future evolution of road geometry and map topology. Future work will incorporate these dynamics to build a more comprehensive world model.

\clearpage  

%
%
\bibliographystyle{splncs04}
\bibliography{main}
\end{document}